\documentclass[letterpaper, 10pt, conference]{ieeeconf}
\IEEEoverridecommandlockouts
\usepackage{pifont}
\usepackage{booktabs}

\usepackage[accsupp]{axessibility}  
\usepackage{tikz}
\usepackage{cite}
\usepackage{amsmath,amssymb,amsfonts}
\usepackage{algorithmic}
\usepackage{graphicx}
\usepackage{textcomp}
\usepackage{makecell}
\usepackage{xcolor}
\usepackage{booktabs} 
\usepackage{multirow}
\usepackage{algorithm}
\usepackage{algorithmic}

\usepackage{hyperref}

\usepackage{orcidlink}
\begin{document}

\title{\LARGE \bf ETA-VLA: Efficient Token Adaptation via Temporal Fusion and Intra-LLM Sparsification for Vision-Language-Action Models} 


\author{
    Yiru Wang$^{1}$,
    Anqing Jiang$^{1}$,
    Shuo Wang$^{1}$, 
    Yuwen Heng$^{1}$, 
    Zichong Gu$^{2}$,
    Hao Sun$^{1}$
    \thanks{
    This work was supported by Bosch Corporate Research. (\textit{Corresponding author: : hao.sun4@cn.bosch.com})} 
    \thanks{$^{1}$
            Yiru Wang,
            Anqing Jiang,
            Shuo Wang,
            Yuwen Heng,
            Hao Sun are with Bosch Corporate Research, Bosch (China) Investment Ltd., Shanghai, China. }
    \thanks{$^{2}$
        Zichong Gu
        is with School of Communication and Information Engineering, Shanghai University, Shanghai, China
    }
}


\maketitle

\begin{abstract}
The integration of Vision-Language-Action (VLA) models into autonomous driving systems offers a unified framework for interpreting complex scenes and executing control commands. However, the necessity to incorporate historical multi-view frames for accurate temporal reasoning imposes a severe computational burden, primarily driven by the quadratic complexity of self-attention mechanisms in Large Language Models (LLMs). To alleviate this bottleneck, we propose \textbf{ETA-VLA}, an \textit{\textbf{E}}fficient \textit{\textbf{T}}oken \textit{\textbf{A}}daptation framework for VLA models. ETA-VLA processes the past $n$ frames of multi-view images and introduces a novel Intra-LLM Sparse Aggregator (ILSA). Drawing inspiration from human driver attention allocation, ILSA dynamically identifies and prunes redundant visual tokens guided by textual queries and temporal consistency. Specifically, we utilize a text-guided scoring mechanism alongside a diversity-preserving sparsification strategy to select a sparse subset of critical tokens, ensuring comprehensive awareness of the driving scene. Extensive experiments on the NAVSIM v2 demonstrate that ETA-VLA achieves driving performance comparable to state-of-the-art baselines while reducing computational FLOPs by approximately 32\%. Notably, our method prunes 85\% of visual tokens and reduces inference FLOPs by 61\%, but still retaining 94\% of the original accuracy on the NAVSIM v2 benchmark.

\end{abstract}

\section{INTRODUCTION}
In autonomous driving, Vision-Language-Action (VLA) models have emerged as a promising direction. By leveraging the commonsense reasoning capabilities of Large Language Models (LLMs), VLAs can effectively map high-dimensional visual inputs and linguistic instructions into executable driving actions. A core objective of VLA research is to enable these models to mimic human driving behavior, which requires not only accurate decision-making but also an efficient allocation of cognitive resources.

Despite their potential, deploying VLA models in real-world driving environments faces a critical challenge: the efficiency-accuracy trade-off in processing high-dimensional spatiotemporal data. Driving is inherently a dynamic process; accurate trajectory prediction requires an understanding of motion history, often necessitating the processing of the past $n$ frames. Standard VLA architectures typically flatten visual tokens from multi-view history and concatenate them with text prompts. As the number of frames $n$ increases, the sequence length grows linearly, leading to a \textit{quadratic explosion} in FLOPs due to the self-attention mechanism ($O(L^2)$). This challenge is further exacerbated in autonomous driving, where each frame contains multiple camera views. Consequently, the total number of tokens—and thus the computational complexity—scales multiplicatively with both the number of frames and the number of views, making the ``token bloat'' problem even more severe and rendering infeasible on vehicle-embedded hardware.

Prior approaches to mitigate this issue generally fall into two categories. (1) \textit{Temporal feature aggregation}. Classic pipelines rely on 3D convolutional networks (C3D)~\cite{tran2015c3d} or recurrent modules to encode spatiotemporal features, while modern transformer-based systems such as ST-P3~\cite{hu2022stp3} and UniAD~\cite{hu2023uniad} fuse temporal information through dedicated perception-prediction modules. (2) \textit{Visual token pruning and merging}. Works like DynamicViT~\cite{rao2021dynamicvit} and ToMe~\cite{bolya2022tome} reduce token redundancy in vision transformers, but are not aware of textual semantics. For large vision-language models (VLMs), SparseVLM~\cite{zhang2025sparsevlm}, and MADTP~\cite{cao2024madtp} leverage cross-modal alignment or textual instructions to guide visual token pruning, yet they either operate outside the LLM or treat temporal frames as independent images, without explicitly modeling multi-view, multi-frame temporal dependencies within the LLM.

In this paper, we propose \textbf{ETA-VLA}, an Efficient Temporal Aggregation framework for VLA models. Our key insight is that efficiency must be addressed at both the temporal and spatial levels: historical observations contain significant redundancy, and even after temporal compression, the resulting spatial representation remains computationally intensive for the LLM. We introduce two complementary components: a Temporal Fusion Module (TFM) and an Intra-LLM Sparse Aggregator (ILSA).
\begin{itemize}
    \item \textbf{Efficient Temporal Fusion:} To mitigate the linear growth of token count with time, TFM operates on the visual encoder features before the LLM. TFM adaptively fuses multi-frame historical features into a concise representation via transformer and learnable weights, effectively compressing the temporal dimension while preserving critical motion cues.
    
    \item \textbf{RoPE-free Semantic Scoring:} Inside the LLM, ILSA performs fine-grained token sparsification. We propose a RoPE-free Semantic Scoring mechanism that computes without the distance-induced bias of Rotary Position Embeddings (RoPE). This ensures visual tokens are scored purely based on their semantic relevance to the driving instruction.
    
    \item \textbf{Diversity-Preserving Sparsification:} To ensure safety, ILSA mimics human attention allocation by employing a per-view recycling strategy. This approach balances global pruning of task-relevant regions with the retention of top-scored tokens in each view, preserving intra-view diversity and preventing the loss of awareness for critical spatial regions like blind spots.
\end{itemize}

\begin{figure*}[htbp!] %
    \centering
    \includegraphics[width=7in]{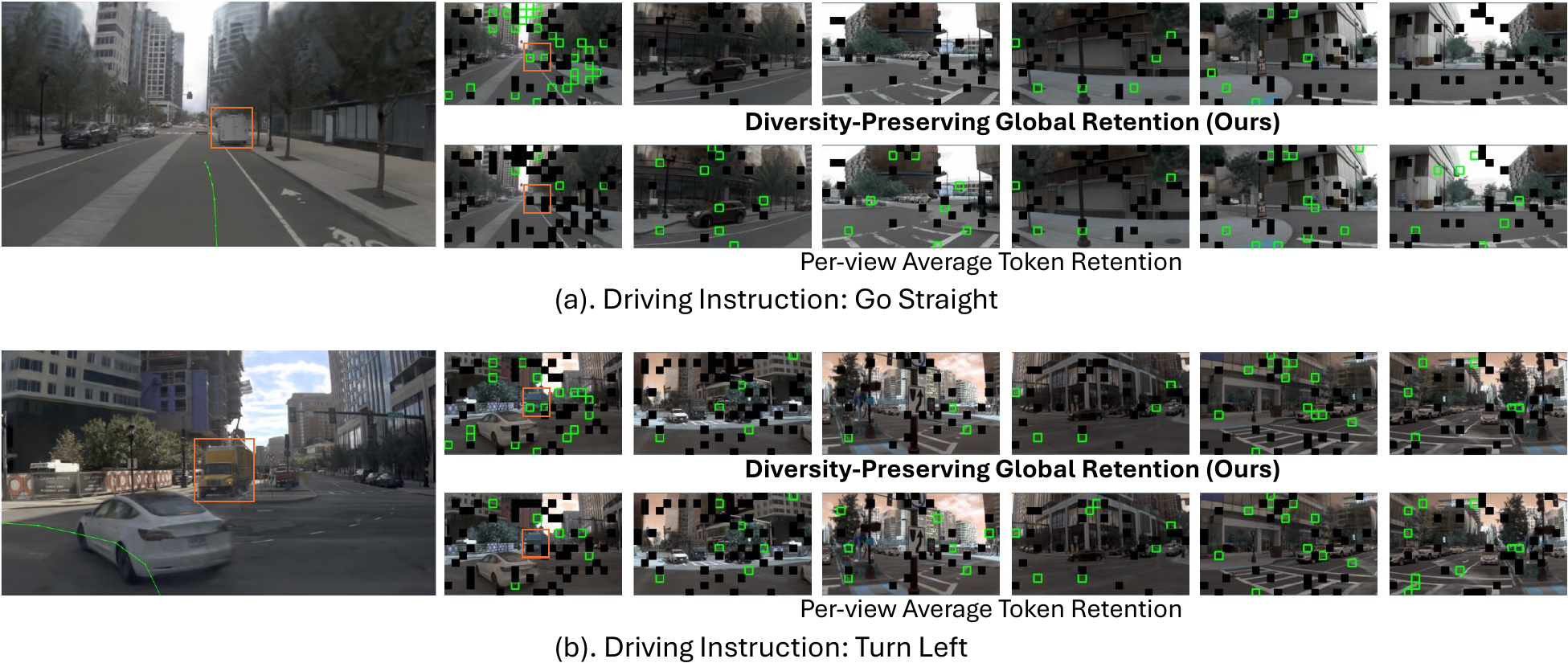} %
    \caption{Visualization of ILSA's token pruning and recycling: The left column shows the front-view camera input with the predicted trajectory (green line) and safety-critical objects (orange boxes). The right columns display multi-view images (ordered as Front, Front Left, Rear Left, Front Right, Rear Right, and Rear) with pruning masks (black boxes), recycled tokens (green boxes), and safety-critical objects (orange boxes). Our method (top) prioritizes critical regions and maintains diversity, mimicking human attention allocation. In contrast, the per-view average method (bottom) dilutes focus.}
    \label{fig:vis} %
\end{figure*}
Fig.~\ref{fig:vis} visualizes the token retention behavior of our ILSA module in two driving scenarios, contrasting our global diversity-preserving strategy with a fixed per-view average retention approach. In the straight-driving scenario in Fig.~\ref{fig:vis}(a), our method prioritizes semantically critical regions in the front view, pruning redundant background. This mirrors human drivers, who focus primarily on the road ahead , when the driving instruction is `go straight'. The per-view average strategy, however, dilutes attention by retaining non-critical tokens. In Fig.~\ref{fig:vis}(b), a human driver must monitor the front for the turn while staying aware of lateral views. Our strategy intelligently balances viewpoint importance, retaining diverse, critical information globally. This aligns with human behavior, which maintains comprehensive awareness during a turn.

Our ILSA module’s global diversity-preserving mechanism mimics human attention allocation, dynamically allocating attention based on the scene. This results in a superior balance of efficiency and performance, making our model more robust for complex driving tasks.

We instantiate ETA-VLA on the NAVSIM v2 benchmark~\cite{dauner2024navsim} and achieve an EPDMS of 85.0 on Navtest while saving 32\% FLOPs, demonstrating its high-fidelity driving performance compared with strong baselines. 
\section{RELATED WORK}
\subsection{Vision-Language-Action Models}
Inspired by integrating LLMs advances in embodied intelligence~\cite{zitkovich2023rt}, recent works integrate multimodal perception and language into unified architectures to bridge the semantic-control gap. These VLA models forms cohesive perception-reasoning-control pipelines that support adaptation to rare scenarios, robustness under partial observability, and comprehension of high-level instructions~\cite{huang2025drivegpt, jiang2025diffvla}. However, they typically process visual inputs frame-by-frame, lacking efficient aggregation of long multi-view temporal history within the LLM, and consequently fail to mimic the dynamic attention allocation characteristic of human drivers.
\subsection{Visual Token Sparsification in VLMs}
\textbf{Vision-only token compression and pruning.}
Works like BLIP-2~\cite{li2023blip2} compress tokens before the LLM. While effective for static images, these fixed schemes discard fine-grained spatial details crucial for driving. Methods like DynamicViT~\cite{rao2021dynamicvit} and ToMe~\cite{bolya2022tome} reduce redundancy in single-modal ViTs but ignore text guidance. HoloV~\cite{zou2025holov} operates on the ViT and rethinks retention holistically; by adaptively distributing pruning budgets across spatial crops, it ensures tokens capture global context rather than isolated features, achieving superior efficiency-accuracy trade-offs. 

\textbf{Text-guided dynamic pruning for VLMs.}
MADTP~\cite{cao2024madtp} leverages vision-language alignment to decide token retention during inference. SparseVLM~\cite{zhang2025sparsevlm} designs a training-free framework using text tokens as “raters” to score visual tokens via self-attention matrices. However, it relies on the standard causal attention matrix, which is influenced by positional encoding and may not fully capture pure semantic relevance, thus failing to achieve the human-like, context-aware attention allocation required for safe driving.

Despite this progress, most methods operate within ViT or use fixed schedules. By contrast, our ETA-VLA performs layer-adaptive sparse processing directly inside LLM, with a RoPE-free semantic scoring mechanism tailored to multi-view, multi-frame driving scenarios, which is not considered by prior works. This design allows our model to mimic human cognitive prioritization, dynamically focusing on task-relevant regions while maintaining awareness of the broader context.
\subsection{Temporal Modeling in Autonomous Driving}
\textbf{Spatial-Temporal Perception in BEV.}
Modern perception systems shift towards Bird's-Eye-View (BEV) representations to aggregate temporal information. BEVFormer~\cite{li2024bevformer} utilizes temporal self-attention to fuse historical features, enhancing detection, while StreamPETR~\cite{wang2023streampetr} introduces a memory queue for efficient long-term context propagation. These demonstrate that explicit temporal fusion is crucial for handling dynamic scenes.

\textbf{Temporal Planning in End-to-End (E2E) System.}
Recent E2E works leverage Transformers to unify tasks with historical context. UniAD~\cite{hu2023uniad} tracks historical agents via query propagation, while VAD~\cite{jiang2023vad} models vectorized elements for efficient trajectory planning. GenAD~\cite{wang2024genad} generalizes temporal learning into a generative framework, highlighting unified reasoning.

However, existing methods typically treat temporal modeling as perception-level aggregation or simple frame concatenation, ignoring semantic reasoning or attention allocation. In contrast, our ETA-VLA employs a Temporal Fusion Module (TFM) to compress redundancy and an Intra-LLM Sparse Aggregator (ILSA) for semantic token selection inside the LLM. This design bridges low-level perception and high-level reasoning, ensuring efficiency and safety by mimicking the way humans process and prioritize temporal and spatial information.

\begin{figure*}[htbp!] %
    \centering
    \includegraphics[width=7in]{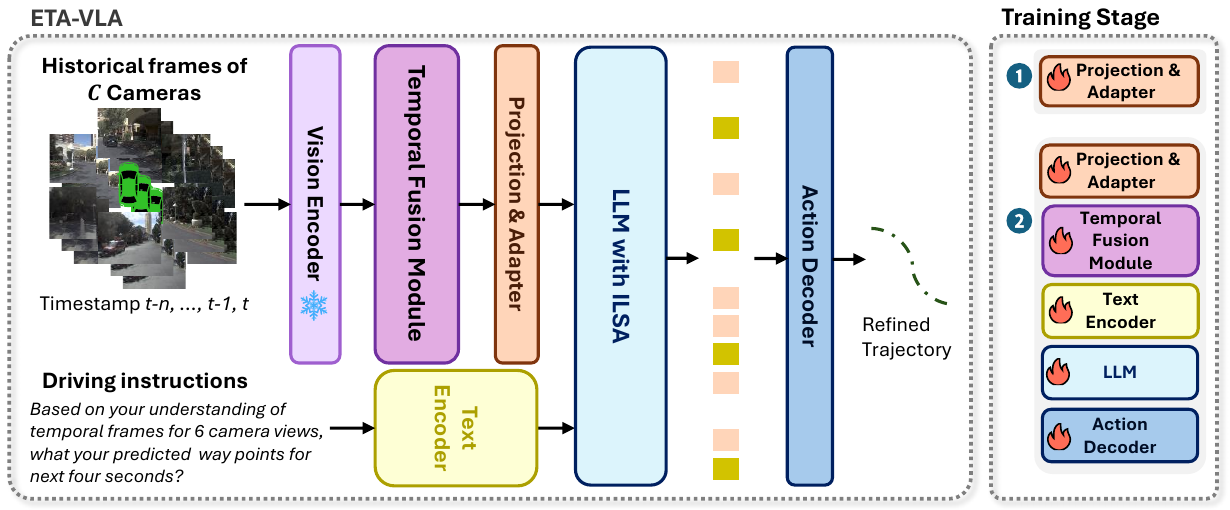} %
    \caption{ETA-VLA Framework with Temporal Fusion Module (TFM),  Intra-LLM Sparse Aggregator (ILSA) and training stage.}
    \label{fig:overview} %
\end{figure*}
\section{METHODOLOGY}
\subsection{System Overview}
The architecture of ETA-VLA is depicted in Fig. \ref{fig:overview}. It comprises three core components: (1) A TFM-based multi-modal encoder that extracts features from historical multi-view images and driving instruction; (2) An LLM backbone augmented with the Intra-LLM Sparse Aggregator (ILSA) to perform dynamic, context-aware token selection, mimicking the selective attention process of human cognition and achieve sparse yet effective reasoning; (3) Action Decoder to decode the sparse tokens from LLM to trajectory and further refined by a pretrained optimizer.

Given a dense token sequence of $n$ historical frames from $C$ cameras as $\mathcal{I}_{\text{dense}}$, the system aims to predict the future trajectory $a$. Unlike standard VLAs that process the full dense sequence, ETA-VLA employs ILSA at selected layers of the LLM to dynamically prune the sequence $\mathcal{I}_{\text{dense}} \to \mathcal{I}_{\text{sparse}}$, thereby reducing the computational cost of subsequent attention operations.
\vspace{-3pt}
\subsection{Temporal Fusion Module (TFM)}
\begin{figure*}[htbp!] %
    \centering
    \includegraphics[width=7in]{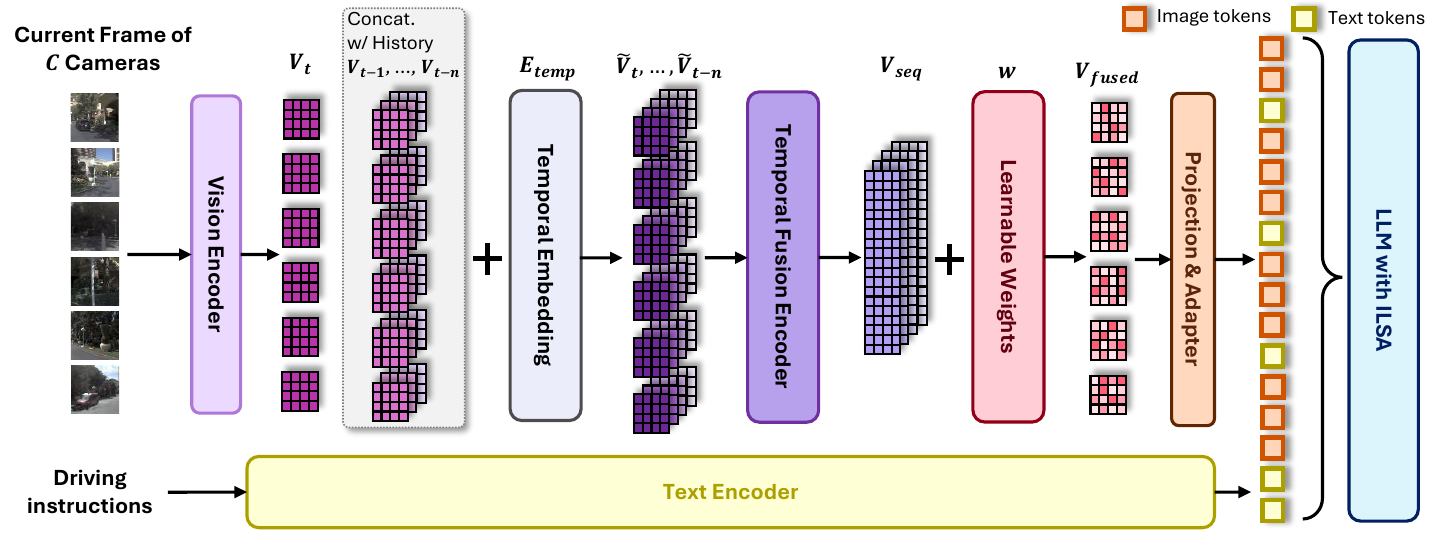} %
    \caption{The Temporal Fusion Module (TFM) for adaptive historical multi-view information aggregation.}
    \label{fig:TFM} %
\end{figure*}
The details of TFM is shown in Fig.~\ref{fig:TFM}. Given a sequence of historical frames ${I_{t-n}, ..., I_{t-1}, I_{t}}$, a pre-trained vision encoder extracts visual features ${\mathbf{V}_{t-n}, ..., \mathbf{V}_t}$, where each $\mathbf{V}_t$ consists of $T$ tokens. Directly concatenating these tokens results in a sequence length of $T \times (n+1)$, which is computationally prohibitive for the subsequent LLM.
To address this, we propose a Transformer-based Temporal Fusion Module. Specifically, we first inject temporal information into the visual features using a learnable time embedding $\mathbf{E}_{\text{temp}} \in \mathbb{R}^{(n+1) \times d}$, where $n+1$ is the number of historical frames plus current frame and $d$ is the feature dimension. The time embedding is added to the visual features to form a time-aware representation:
\begin{equation}
\tilde{\mathbf{V}}_t = \mathbf{V}_t + \mathbf{E}_{\text{temp}}(t),
\end{equation}
where $\mathbf{E}_{\text{temp}}(t)$ is the embedding for the $t$-th frame.
Next, we pass the time-aware features ($\tilde{\mathbf{V}}_{t-n}, \dots, \tilde{\mathbf{V}}_{t}$) through a Temporal Fusion Encoder to capture temporal dependencies $\mathbf{V}_{\text{seq}}$ via self-attention. Finally, we perform time-weighted aggregation to compress the temporal dimension. Instead of fixed linear weights, we use learnable time weights $\mathbf{w} \in \mathbb{R}^{(n+1)}$. These weights are applied to the temporal tokens, and the weighted sum is normalized to produce a compact visual feature map:
\begin{equation}
\mathbf{V}_{\text{fused}} = \frac{\sum_{i=t-n}^{t} w_i \cdot \mathbf{V}_{\text{seq}}^i}{\sum_{i=t-n}^{t} w_i}.
\end{equation}
This mechanism adaptively aggregates relevant information across time steps, producing a single, compact visual feature map $\mathbf{V}_{\text{fused}}$ that encapsulates the temporal context. This fused representation is then projected and passed through an adapter~\cite{jiang2024senna}, and then concatenated with text tokens to form the input sequence for the LLM.

\subsection{Intra-LLM Sparse Aggregator (ILSA)}
\subsubsection{ILSA Overview}
\begin{figure*}[htbp!] 
\centering
\includegraphics[width=7in]{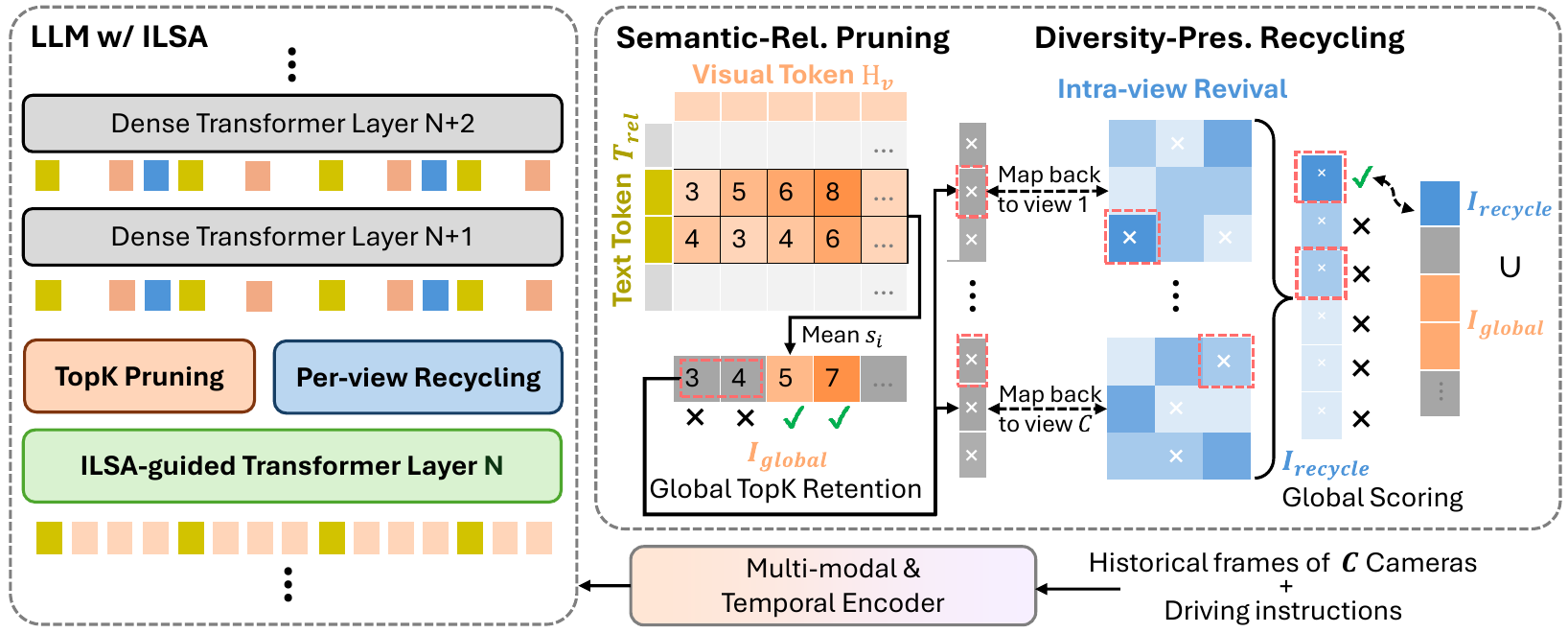} 
\caption{Intra-LLM Sparse Aggregator (ILSA) Framework with Semantic-Relative Pruning and Diversity-Preserving Recycling.}
\label{fig:ilsa} 
\end{figure*}
The core of ETA-VLA lies in the ILSA module as illustrated in Fig. ~\ref{fig:ilsa}, which is deeply integrated into the LLaMA backbone. Unlike off-the-shelf sparse attention mechanisms that rely on external scoring networks, ILSA utilizes the intrinsic self-attention logits of the LLM to guide token pruning. To balance computational efficiency with the need for explicit attention weights, we design a Hybrid Execution Strategy within the attention forward pass.
\subsubsection{Hybrid Execution Strategy}
Our Hybrid Execution Strategy is centered on the efficient processing of sparse layers, which are the primary drivers of our model’s efficiency gains. To access the attention weights required for our novel token selection process, we employ an eager attention implementation on these designated sparse layers. For all other layers, we retain the highly optimized, memory-efficient Scaled Dot-Product Attention (SDPA)/FlashAttention. This targeted approach introduces only a minor overhead on a small fraction of layers, which is significantly outweighed by the substantial reduction in sequence length and FLOPs achieved through token pruning.

\subsubsection{Dynamic Text Anchor Selection}
Existing methods typically calculate visual importance using all text tokens, including padding or irrelevant instructions, which introduces noise. To address this, we propose a dynamic anchor selection mechanism. While sharing the intuition of selecting relevant linguistic ``raters'' with SparseVLM~\cite{zhang2025sparsevlm}, our approach introduces two fundamental distinctions tailored for autonomous driving tasks. 

Let $\mathbf{H} \in \mathbb{R}^{L \times D}$ denote the input hidden states of the \textit{current layer}. We extract the corresponding subsets $\mathbf{H}_{\mathcal{V}}$ and $\mathbf{H}_{\mathcal{T}}$ via indexing, where $\mathcal{V}$ and $\mathcal{T}$ identify visual and text positions, respectively. Before the attention calculation, we compute the semantic similarity:
\begin{equation}
\mathbf{S} = \text{softmax}(\mathbf{H}_{\mathcal{V}} \mathbf{H}_{\mathcal{T}}^T).
\end{equation}
We then calculate an alignment score $\tau$ for each text token by averaging its similarity across all visual tokens. Text tokens whose scores exceed $\tau$ (set as the mean score) are selected as relevant anchors $\mathcal{T}_{rel}$:
\begin{equation}
\mathcal{T}_{rel} = \left\{ t \in \mathcal{T} \mid \frac{1}{|\mathcal{V}|} \sum_{v \in \mathcal{V}} \mathbf{S}_{v,t} > \tau \right\}.
\end{equation}
This allows the model to adaptively refine linguistic guidance as cross-modal semantics evolve through successive transformer blocks. 

Second, whereas SparseVLM is primarily designed for sequential vision-language inputs, our \textit{indexing-based extraction} ($\mathcal{V}, \mathcal{T}$) is specifically optimized for interleaved multi-view sequences. This ensures that task-relevant guidance is precisely localized even when visual tokens from multiple cameras are complexly interspersed with instructions, fulfilling a critical structural requirement for safe and accurate trajectory prediction.

\subsubsection{RoPE-free Semantic Scoring}
A naive approach to scoring visual tokens involves using the standard causal attention weights. However, LLMs utilizing Rotary Position Embeddings (RoPE), the attention score is heavily biased by relative distance. This is detrimental for driving scenarios where relevant cues may appear distant in the sequence.
We propose RoPE-free Semantic Scoring. We first obtain the raw query $\mathbf{Q}_{raw}$ and key $\mathbf{K}_{raw}$ states via projection $\text{Proj}_Q(\mathbf{H})$ and $\text{Proj}_K(\mathbf{H})$, respectively. We then compute the pruning attention logits, without positional rotation. Here, $d$ denotes the dimension of each attention head, used to scale the dot-product attention:
\begin{equation}
\mathbf{A}_{\text{prune}} = \text{softmax}\left( \frac{\mathbf{Q}_{raw} \mathbf{K}_{raw}^T}{\sqrt{d}} \right).
\end{equation}
Crucially, we remove the causal mask for $\mathbf{A}_{\text{prune}}$, allowing bidirectional semantic flow. This matrix purely reflects the semantic relevance between the instruction and the visual tokens.
\subsubsection{Diversity-Preserving Token Selection}
Utilizing the semantic attention matrix $\mathbf{A}_{\text{prune}}$ and the selected text anchors $\mathcal{T}_{rel}$, we derive the importance score $s_i$ for each visual token by averaging the attention weights it receives from $\mathcal{T}_{rel}$:
\begin{equation}
s_i = \frac{1}{|\mathcal{T}_{rel}|} \sum_{t \in \mathcal{T}_{rel}} (\mathbf{A}_{\text{prune}})_{t,i}
\end{equation}
We propose a two-stage selection strategy to balance efficiency and representation completeness:
\paragraph{Global Semantic-Relative Pruning:}
We perform a global Top-$K$ selection across all visual tokens by ranking $s_i$ to identify the most task-relevant tokens $\mathcal{I}_{\text{global}}$. While this focuses on instruction-relevant regions, it risks discarding tokens that are semantically less relevant but spatially unique.
\paragraph{Diversity-Preserving Recycling:}
To recover discarded tokens that possess high feature diversity, we compute a diversity score $d_i$ for tokens not in $\mathcal{I}_{\text{global}}$ based on their feature distinctiveness within their respective views:
\begin{equation}
d_i = 1 - \frac{1}{N-1} \sum_{\substack{j=1 \\ j \neq i}}^{N} \frac{\mathbf{h}_i \cdot \mathbf{h}_j}{\|\mathbf{h}_i\|_2 \|\mathbf{h}_j\|_2},
\end{equation}
where $\mathbf{h}_i$ is the feature vector of the $i$-th token and $N$ is the total number of tokens of each view. We then select the top-$k$ tokens $\mathcal{I}_{\text{recycle}}$ with the highest diversity scores globally.
The final set of retained tokens $\mathcal{I}_{\text{final}}$ ($K+k$ tokens) is the union of both sets:
\begin{equation}
\mathcal{I}_{\text{final}} = \mathcal{I}_{\text{global}} \cup \mathcal{I}_{\text{recycle}}.
\end{equation}
This strategy supplements the task-relevant tokens with diverse features, enhancing the model's robustness to varying scenes. The whole ILSA algorithm is summarised in Algo.~\ref{alg:forward}.
\begin{algorithm}[t]
\caption{Intra-LLM Sparse Aggregator (ILSA)}
\label{alg:forward}
\scriptsize 
\begin{algorithmic}[1]
\REQUIRE Hidden states $\mathbf{H}$, View IDs $\mathbf{C}_{\text{view}}$, Layer index $N$
\STATE $\triangleright$ \textbf{Stage 0: Dynamic Text Selection}
\STATE $\mathbf{S} \leftarrow \text{Sim}(\mathbf{H}_{\mathcal{V}}, \mathbf{H}_{\mathcal{T}})$
\STATE $\mathcal{T}_{rel} \leftarrow \text{Threshold}(\text{Mean}(\mathbf{S}, \text{dim=vis}))$
\STATE $\triangleright$ \textbf{Stage 1: Attention \& Scoring}
\STATE $\mathbf{Q}_{raw}, \mathbf{K}_{raw}, \mathbf{V} \leftarrow \text{Proj}(\mathbf{H})$
\STATE $\mathbf{Q}_{rot}, \mathbf{K}_{rot} \leftarrow \text{RoPE}(\mathbf{Q}_{raw}, \mathbf{K}_{raw})$
\STATE $\mathbf{O} \leftarrow \text{Attn}(\mathbf{Q}_{rot}, \mathbf{K}_{rot}, \mathbf{V})$ \COMMENT{Main Forward with Causal Mask}
\STATE $\mathbf{A}_{\text{prune}} \leftarrow \text{Attn}(\mathbf{Q}_{raw}, \mathbf{K}_{raw}, \mathbf{V})$ \COMMENT{RoPE-free Weights, No Mask}
\STATE $s_i \leftarrow \text{Mean}_{t \in \mathcal{T}_{rel}}((\mathbf{A}_{\text{prune}})_{t,i}), \forall i \in \mathcal{V}$ \COMMENT{Text-guided Saliency}
\STATE $\triangleright$ \textbf{Stage 2: Pruning \& Recycling}
\STATE $\mathcal{I}_{\text{global}} \leftarrow \text{TopK}(s, K)$
\STATE $\mathbf{d} \leftarrow \text{Diversity}(\mathbf{H}_{\mathcal{V} \setminus \mathcal{I}_{\text{global}}})$
\STATE $\mathcal{I}_{\text{recycle}} \leftarrow \text{TopK}(\mathbf{d}, k_{\text{recycle}})$
\STATE $\mathcal{I}_{\text{final}} \leftarrow \mathcal{I}_{\text{global}} \cup \mathcal{I}_{\text{recycle}}$
\STATE \textbf{return} $\mathbf{O}, \mathcal{I}_{\text{final}}$
\end{algorithmic}
\end{algorithm}
\subsubsection{Computational Complexity Analysis}
The computational complexity of standard self-attention is $O(L^2)$, where $L$ is the sequence length. This quadratic growth becomes prohibitive as the number of frames $n$ increases, and the complexity is further amplified in the autonomous driving domain where each frame contains multiple camera views. In contrast, the ILSA module significantly reduces this complexity.

By adopting ILSA to select a fixed number of tokens ($K+k$), the sequence length for the main attention computation is reduced from $L$ to a constant. This lowers the dominant self-attention complexity from $O(L^2)$ to $O((K+k)^2)$. Furthermore, since the sequence length is reduced for the current layer, all subsequent layers also operate on this shorter sequence, avoiding the quadratic complexity $O(L^2)$ for every following layer. This results in a cascading efficiency gain. Therefore, the overall complexity of the ILSA-augmented layer is dominated by linear-time operations, making it effectively $O(L)$ with respect to the original sequence length $L$. This represents a substantial reduction in FLOPs, making it feasible for VLA models in autonomous driving.

\subsection{Action Decoding and Training}
\subsubsection{Action Decoder and Optimization}
Instead of directly auto-regressively generating discrete actions via the LLM, we employ a lightweight Transformer Decoder to predict continuous future trajectories. Specifically, the LLM processes the sparsified visual sequence $\mathcal{I}_{\text{sparse}}$ and outputs hidden states $\mathcal{H}$. These states are projected as memory features for the decoder. We introduce a set of learnable temporal embeddings $E_{\text{traj}} \in \mathbb{R}^{T_r \times m}$ as queries, where $T_r$ is the trajectory length and $m$ is the embedding length. The embeddings attend to the memory features via cross-attention to aggregate spatial-temporal context. The decoder output is finally mapped to trajectory of length $T_r$ through a linear layer. 

Then a lightweight pre-trained Scorer module, introduced in ~\cite{jiang2025irl} and trained on NAVSIM v2~\cite{dauner2024navsim}, optimized the output trajectory to a refined trajectory by predicting the Extended Predictive Driver Model Score (EPDMS). This score assesses trajectories based on safety, efficiency, and comfort criteria, enabling simulator-free evaluation.

To ensure accurate and physically plausible trajectory generation, the model is supervised by a composite loss function:
\begin{equation}
\mathcal{L}_{\text{total}} = \mathcal{L}_{\text{traj}} + \lambda_{1}\mathcal{L}_{x} + \lambda_{2}\mathcal{L}_{\text{vel}} + \lambda_{3}\mathcal{L}_{\text{end}},
\end{equation}
where $\mathcal{L}_{\text{traj}}$ denotes the Smooth L1 loss between predicted and ground-truth coordinates. To enforce physical feasibility, we introduce a velocity smoothness loss $\mathcal{L}_{\text{vel}}$ that minimizes the difference between predicted and ground-truth velocities. Additionally, we apply a specialized loss $\mathcal{L}_{x}$ with a higher weight on lateral coordinates to enhance steering stability, and an endpoint loss $\mathcal{L}_{\text{end}}$ to improve final destination accuracy. The weights $\lambda_{1}, \lambda_{2}, \lambda_{3}$ are used to balance these objectives.

\subsubsection{Training Strategy}
\label{sec:training} 
Our training process consists of three sequential stages, as illustrated in Fig~\ref{fig:overview}: First, we pre-train the projection and adapter. Then, we jointly optimize the multi-modal, temporal encoder, LLM and the trajectory decoder by minimizing $\mathcal{L}_{\text{total}}$. 
\section{EXPERIMENTS}
To validate the effectiveness of our proposed framework, we conduct comprehensive evaluations on the established NAVSIM v2 benchmark~\cite{dauner2024navsim}, comparing against a range of state-of-the-art methods. We further perform detailed ablation studies to isolate the contribution of each key component within our architecture.
\subsection{Experimental Setup and Metrics}
We evaluate on the NAVSIM v2 Navtest (open-loop) and Navhard (pseudo closed-loop with perturbations) benchmarks. The primary metric, EPDMS, aggregates nine sub-scores: No At-Fault Collision (NC), Drivable Area Compliance (DAC), Driving Direction Compliance (DDC), Traffic Light Compliance (TLC), Ego Progress (EP), Time to Collision (TTC), Lane Keeping (LK), history Comfort (HC), and Extended Comfort (EC). The EPDMS score provides a holistic measure of an autonomous driving system's safety, efficiency, and comfort.

Built upon LLaVA-v1.5-7B~\cite{liu2024improved} with a ViT-L/14 visual encoder and Vicuna-7B LLM, our model is trained in two stages: (1) pre-training on the trainval split for 1 epoch using 8$\times$A800 GPUs (batch size 32, lr $1\times10^{-3}$); (2) fine-tuning on navtrain for 1 epoch (8$\times$A800, batch size 8, lr $1\times10^{-5}$). During the training phase, we use the preceding 2 frames as historical information at each timestep.

\subsection{Quantitative Results}
\subsubsection{Performance on Navtest Benchmark}
Table~\ref{tab:vlm_exp} summarizes the experimental results on the NAVSIM v2 Navtest benchmark. Notably, our ETA-VLA model (with 35\% token pruning) establishes a new state-of-the-art with an EPDMS score of 85.0, significantly outperforming established baselines such as VADv2 (76.6), Transfuser (76.7), and prior SOTA methods including the diffusion-based DiffusionDrive (84.3) and GTRS w/ SimScale (84.6). Remarkably, this score closely approaches the human expert performance of 90.3, demonstrating the model’s human-like decision-making capabilities. Such robust performance is anchored by exceptional results in NC (99.2) and HC (98.3) scores. These results highlight the effectiveness of our efficient Temporal Fusion in maintaining stable temporal reasoning and the Diversity-Preserving Sparsification strategy in ensuring comprehensive spatial awareness, which are crucial for safe and compliant driving.
\begin{table*}[htbp!]
  \centering
  \small
  \caption{Evaluation on NAVSIM v2 Navtest Benchmark for Planning Oriented Autonomous Driving.}
  \begin{tabular}{@{}l|ccccccccc|c@{}}
    \toprule
    \textbf{Method} & \textbf{NC $\uparrow$} & \textbf{DAC $\uparrow$} & \textbf{DDC $\uparrow$} & \textbf{TL $\uparrow$} & \textbf{EP $\uparrow$} & \textbf{TTC $\uparrow$} & \textbf{LK $\uparrow$} & \textbf{HC $\uparrow$} & \textbf{EC $\uparrow$} & \textbf{EPDMS $\uparrow$}\\
    \midrule
    Human Agent &100 &100 &99.8 &100 &87.4 &100 &100 &98.1 &90.1 &90.3 \\
    \midrule
    Transfuser\cite{chitta2022transfuser} &96.9 &89.9 &97.8 &99.7 &87.1 &95.4 &92.7 &98.3 &87.2 &76.7\\
    VADv2\cite{chen2024vadv2} &97.3 &91.7 &77.6 &92.7 &100 &99.9 &98.2 &66.0 &97.4 &76.6\\
    HydraMDP++\cite{lihydra} &97.2 &97.5 &99.4 &99.6 &83.1 &96.5 &94.4 &98.2 &70.9 &81.4\\
    GTRS\cite{li2025generalized} &97.6 &98.5 &99.5 &99.9 &89.5 &97.2 &96.8 &97.2 &57.2 &84.0\\
    GTRS w/ SimScale\cite{tian2025simscale} &98.4 &98.8 &99.4 &99.9 &87.9 &98.1 &96.4 &97.6 &58.8 &84.6\\
    DiffusionDrive\cite{liao2025diffusiondrive} &98.0 &96.0 &99.5 &99.8 &87.7 &97.1 &97.2 &98.3 &87.6 &84.3\\
    \textbf{ETA-VLA (Ours)} &99.2 &96.7 &99.3 &99.9 &87.2 &98.4 &97.8 &98.3 &65.1 &\textbf{85.0}\\
    \bottomrule
  \end{tabular}
  \label{tab:vlm_exp}
\end{table*}

\begin{table*}[htbp!]
  \centering
  \small
  \caption{Evaluation on NAVSIM-V2 Navhard Benchmark for Planning Oriented Autonomous Driving.}
  \centering
  \begin{tabular}{l|l| l l l l l l l l l |l}
    \toprule
    \textbf{Method} 
    & \textbf{Stage}
    & $\textbf{NC$\uparrow$}$
    & $\textbf{DAC$\uparrow$}$
    & $\textbf{DDC$\uparrow$}$
    & $\textbf{TLC$\uparrow$}$
    & $\textbf{EP$\uparrow$}$
    & $\textbf{TTC$\uparrow$}$
    & $\textbf{LK$\uparrow$}$ 
    & $\textbf{HC$\uparrow$}$ 
    & $\textbf{EC$\uparrow$}$ 
    & $\textbf{EPDMS$\uparrow$}$  \\
    \midrule
    
    PDM-Closed~\cite{dauner2023parting} & \makecell{Stage 1 \\ Stage 2} & \makecell{94.4 \\ 88.1} & \makecell{98.8 \\ 90.6} & \makecell{100 \\ 96.3} & \makecell{99.5 \\ 98.5} & \makecell{100 \\ 100} & \makecell{93.5 \\ 83.1 } & \makecell{99.3 \\ 73.7} & \makecell{87.7 \\ 91.5} & \makecell{36.0 \\ 25.4} & 51.3    \\
    \midrule
    Transfuser~\cite{chitta2022transfuser} &  \makecell{Stage 1 \\ Stage 2} &
    \makecell{96.2 \\ 77.7} & \makecell{79.5 \\ 70.2} & \makecell{99.1 \\ 84.2} & \makecell{99.5 \\ 98.0} & \makecell{84.1 \\ 85.1} & \makecell{95.1 \\ 75.6} & \makecell{94.2 \\ 45.4} & \makecell{97.5 \\ 95.7} & \makecell{79.1 \\ 75.9} & 23.1    \\
    \midrule

    DiffusionDrive~\cite{liao2025diffusiondrive} & \makecell{Stage 1 \\ Stage 2}  &
    \makecell{95.9 \\ 79.5} & \makecell{84.0 \\ 72.8} & \makecell{98.6 \\ 84.1} & \makecell{99.8 \\ 98.4} & \makecell{84.4 \\ 87.5} & \makecell{96.0 \\ 76.2} & \makecell{95.1 \\ 46.6} & \makecell{97.6 \\ 96.1} & \makecell{71.1 \\ 62.4} &  26.0  \\
    \midrule
   
    Senna-E2E~\cite{jiang2024senna} &   \makecell{Stage 1 \\ Stage 2}  &
    \makecell{95.6 \\ 78.6} & \makecell{86.0 \\ 74.8} & \makecell{98.9 \\ 84.8} & \makecell{99.6 \\ 98.2} & \makecell{83.9 \\ 88.2} & \makecell{95.1 \\ 75.7} & \makecell{95.3 \\ 46.9} & \makecell{97.6 \\ 96.0} & \makecell{75.6 \\ 65.8} & 27.2    \\
    \midrule

    GTRS w/ SimScale\cite{tian2025simscale} & \makecell{Stage 1 \\ Stage 2}  &
    \makecell{99.1 \\ 92.3} & \makecell{98.2 \\ 93.8} & \makecell{99.8 \\ 94.9} & \makecell{100 \\ 99.3} & \makecell{71.9 \\ 75.4} & \makecell{99.3 \\ 90.7} & \makecell{95.6 \\ 60.3} & \makecell{93.8 \\ 95.9} & \makecell{28.0 \\ 37.4} &  47.2  \\
    \midrule
        
    \makecell[l]{\textbf{ETA-VLA (Ours)}} & \makecell{Stage 1 \\ Stage 2}  &
    \makecell{98.9 \\ 86.7} & \makecell{98.4 \\ 82.7} & \makecell{99.3 \\ 90.0} & \makecell{99.6 \\ 98.7} & \makecell{85.2 \\ 85.0} & \makecell{99.1 \\ 83.4} & \makecell{99.1 \\ 55.9} & \makecell{97.8 \\ 96.9} & \makecell{73.3 \\ 63.4} &  \textbf{48.0}  \\

  \bottomrule
 \end{tabular}
\label{table:result_navhard}
\end{table*}

\subsubsection{Performance on Navhard Benchmark}
Table~\ref{table:result_navhard} presents the evaluation results on the rigorous Navhard benchmark, which is designed to test robustness in pseudo closed-loop scenarios. Our ETA-VLA model (35\% token pruning rate) achieves a state-of-the-art EPDMS score of 48.0, substantially outperforming established baselines such as DiffusionDrive (26.0) and Transfuser (23.1). Notably, compared to the previous leading method GTRS w/ SimScale (47.2), ETA-VLA demonstrates superior robustness with a clear margin. This advantage is primarily attributed to our proposed architectural innovations. The proposed robust temporal and diversity preserving representation is evidenced by significant improvements in key safety metrics, indicating better handling of driving commands and boundary conditions. Collectively, these results demonstrate the robustness and generalizability of the ETA-VLA architecture across both open-loop (Navtest) and pseudo closed-loop (Navhard) settings.

\subsubsection{Efficiency Analysis of ETA-VLA}
\begin{table}[htbp!]
    \scriptsize
    \centering
    \caption{Efficiency Analysis on Navtest.}
    \label{tab:flops_eval}
    \begin{tabular}{c |c| c |c|c }
        \toprule
         \textbf{ILSA}  &\textbf{Sparse Layers} &\textbf{Pruning Rate} &  \textbf{GFLOPs} $\downarrow$ & \textbf{EPDMS} $\uparrow$ \\
        \midrule
        w/o &/ &/ & 9,105  &  84.8 \\
        \midrule
        w/  &2 &35\% &{6,091} & {84.4}  \\ 
        w/  &2 &60\% &{4,773} & {82.8}  \\ 
        w/  &2 &85\% &{3,556} & {79.8}  \\
        \midrule
        w/  &4 &35\% &\textbf{6,190} & \textbf{85.0}  \\ 
        w/  &4 &60\% &{4,963} & {84.7}  \\ 
        w/  &4 &85\% &{3,830} & {80.5}  \\
        \midrule
        w/  &2, 4 &35\%, 60\% &{5,447} & {78.9}   \\ 
        w/  &2, 4 &60\%, 85\% &{4,068} & {77.3}  \\
        \bottomrule 
    \end{tabular}
\end{table}
\subsection{Ablation Studies}
\begin{table*}[htbp]
  \centering
  \small
  \caption{Ablation study on NAVSIM v2 Navtest Benchmark. }
  \label{tab:vla_exp}

  \begin{tabular}{@{}c|c|ccccccccc|c@{}}
    \toprule
    \makecell{\textbf{Temporal} \\ \textbf{Fusion}} & 
    \begin{tabular}{@{}c@{}}
         \textbf{Pruning \& Recycling} \\
         Ours \hspace{0.5em} Random \hspace{0.5em} HoloV \hspace{0.5em} SparseVLM
    \end{tabular} 
    & 
    NC & DAC & DDC & TL & EP & TTC & LK & HC & EC & EPDMS $\uparrow$\\
    \midrule
    
    \ding{55} & \ding{51} \quad \quad \hspace{4.5em} \hspace{2.5em} \quad \quad \quad  \hspace{1.5em} & 97.3 & 92.1 & 98.7 & 99.8 & 81.9 & 96.3 & 94.6 & 98.3 & 68.7 & 77.2 \\
    \midrule
    \ding{51} & \ding{51}\hspace{6.5em}  & 93.1 & 77.9 & 92.7 & 99.6 & 86.0 & 91.5 & 89.4 & 98.3 & 65.4 & 64.0 \\
    \ding{51} &  \quad \quad \hspace{5.5em} \ding{51} \hspace{0.5em} \quad \quad \quad \hspace{1.5em} &  97.3 & 94.5 & 99.1 & 99.8 & 83.5 & 96.3 & 95.4 & 98.3 & 66.7 & 79.7  \\
    \ding{51} &  \quad \quad \hspace{0.5em}  \hspace{6.5em} \quad \quad \quad \ding{51} &  98.9 & 95.7 & 99.3 & 99.9 & 86.8 & 98.0 & 97.2 & 98.3 & 63.0 & 83.3  \\
    \midrule
    \ding{51} & \ding{51} \quad \quad \hspace{4.5em} \hspace{2.5em} \quad \quad \quad  \hspace{1.5em} & 99.6 & 99.1 & 99.7 & 99.9 & 89.2 & 99.4 & 98.9 & 98.4 & 66.0 & \textbf{85.0} \\
    \bottomrule
  \end{tabular}
\end{table*}
Table~\ref{tab:flops_eval} demonstrates the pivotal role of ILSA in balancing computational efficiency and planning performance. Compared to the dense baseline (9,105 GFLOPs, 84.8 EPDMS), ILSA yields substantial computational reductions across all configurations. We configure the token retention strategy by selecting 90\% of the tokens via top-$K$ scoring, while the remaining 10\% are recycled tokens to ensure diversity. Notably, the model achieves an optimal trade-off when applied at Layer 4 with a 35\% pruning rate: it not only reduces GFLOPs to 6,190 (reducing 32\%) but also outperforms the dense baseline with an EPDMS of 85.0. This improvement indicates that ILSA effectively filters out task-irrelevant noise, acting as a beneficial sparsification module rather than a mere compression technique. As pruning rates increase, performance naturally declines, yet Layer 4 consistently exhibits superior robustness compared to Layer 2. For instance, at a 60\% pruning rate, Layer 4 maintains a high EPDMS of 84.7, whereas Layer 2 drops significantly to 82.8. This suggests that intermediate features at Layer 4 can withstand aggressive sparsification without compromising planning capabilities. However, applying pruning sequentially at both Layer 2 and 4 leads to severe performance degradation (e.g., 78.9 EPDMS), implying that excessive multi-stage sparsification disrupts critical feature continuity. 

We further explored applying ILSA to even deeper layers (e.g., Layer 6 or 8), but observed significant performance drops. We attribute this to the fact that features in these final layers are highly compressed and semantically dense, serving as direct prerequisites for the planning head. Pruning at these critical stages risks removing essential planning primitives, leading to irreversible information loss that cannot be recovered by subsequent layers. In summary, a single sparse layer at an appropriate depth offers the most effective balance between efficiency and performance.

To validate our design choices, we conduct ablation studies on the NAVSIM v2 navtest benchmark (Table~\ref{tab:vla_exp}). We evaluate the impact of two key components: Temporal Fusion, Pruning and Recycling. Disabling Temporal Fusion significantly degrades performance, dropping the EPDMS score to 77.2, indicating its critical role in coherent temporal reasoning. 

For the pruning strategy, we compare our method against random pruning, HoloV~\cite{zou2025holov}, and SparseVLM~\cite{zhang2025sparsevlm} using the same pruning rate as 35\%. As shown in Table~\ref{tab:vla_exp}, random pruning yields the lowest score (64.0), failing to retain critical information for driving. In contrast, our method achieves the best performance (EPDMS 85.0), significantly outperforming HoloV~\cite{zou2025holov} (79.7) and SparseVLM~\cite{zhang2025sparsevlm} (83.3). This validates the superiority of our RoPE-free semantic scoring and diversity-preserving recycling mechanism. Unlike HoloV\cite{zou2025holov}, which relies on fixed spatial partitioning and hinders fine-grained semantic capture in complex driving scenes—or SparseVLM~\cite{zhang2025sparsevlm}, which remains constrained by positional biases in standard attention matrices and the spatial coordinate blurring inherent in its element-wise sum reconstruction (leading to a 1.5-point decrease), our approach dynamically focuses on task-relevant regions across multi-view inputs to ensure the precision required for trajectory prediction. This mimics human attention allocation, leading to a superior efficiency-accuracy trade-off. 

\section{CONCLUSION}
We introduced ETA-VLA, a novel Vision-Language-Action (VLA) model designed to address the computational challenges of multi-frame, multi-view input in autonomous driving. Our solution comprises two complementary components: a Temporal Fusion Module (TFM) for compressing redundant historical information and an Intra-LLM Sparse Aggregator (ILSA) for dynamic token selection. By combining these, we effectively reduce the computational burden while preserving critical motion cues and semantic relevance. Our experiments on the NAVSIM v2 benchmark demonstrate that ETA-VLA achieves an EPDMS of 85.0 on Navtest while reducing FLOPs by 32\%, proving that it significantly outperforms the dense baseline in both efficiency and performance.

Crucially, our work demonstrates that mimicking human attention allocation—the guiding principle behind both TFM and ILSA—provides a powerful and generalizable solution for balancing efficiency and reasoning in VLA models. The TFM mimics the human ability to summarize past events, while the ILSA mimics the selective attention process, allowing our model to focus computational resources on task-critical regions while maintaining comprehensive scene awareness. This work paves the way for deploying high-reasoning VLA models on resource-constrained automotive hardware.

%
%
\bibliographystyle{splncs04}
\bibliography{main}
\end{document}